%% file: main.tex
\documentclass[11pt, a4paper, onecolumn, copyright]{AweAI}

\usepackage{multirow}
\usepackage[authoryear, round]{natbib} % sort&compress
\usepackage{graphicx}
\usepackage[table,dvipsnames]{xcolor}
\usepackage{microtype}
\usepackage{subcaption}
\usepackage{booktabs}

\usepackage{listings} 
\usepackage{makecell}
\usepackage[most]{tcolorbox}
\tcbuselibrary{listings}
\usepackage[htt]{hyphenat}
\usepackage{xspace}
\usepackage{textcomp}
\usepackage{amsmath}
\usepackage{amssymb}
\usepackage{mathtools}
\usepackage{amsthm}
\usepackage{enumitem}
\usepackage{pifont}
\definecolor{darkgreen}{RGB}{50,100,0}
\definecolor{darkred}{RGB}{200, 0, 0}

\usepackage{xurl}
\usepackage{titletoc}

\definecolor{tableblue}{HTML}{F2F7FF}
\definecolor{headergray}{HTML}{444444}

\colorlet{DarkGreen}{green!50!black}
\colorlet{DarkRed}{red}

\newcommand{\msem}[2]{#1\,\mbox{\scalebox{0.78}{$\pm#2$}}}

\usepackage{xcolor}
\newcommand{\pos}[1]{\textcolor{darkred}{+#1}} % red
\newcommand{\github}{\raisebox{-1.5pt}{\includegraphics[height=1.05em]{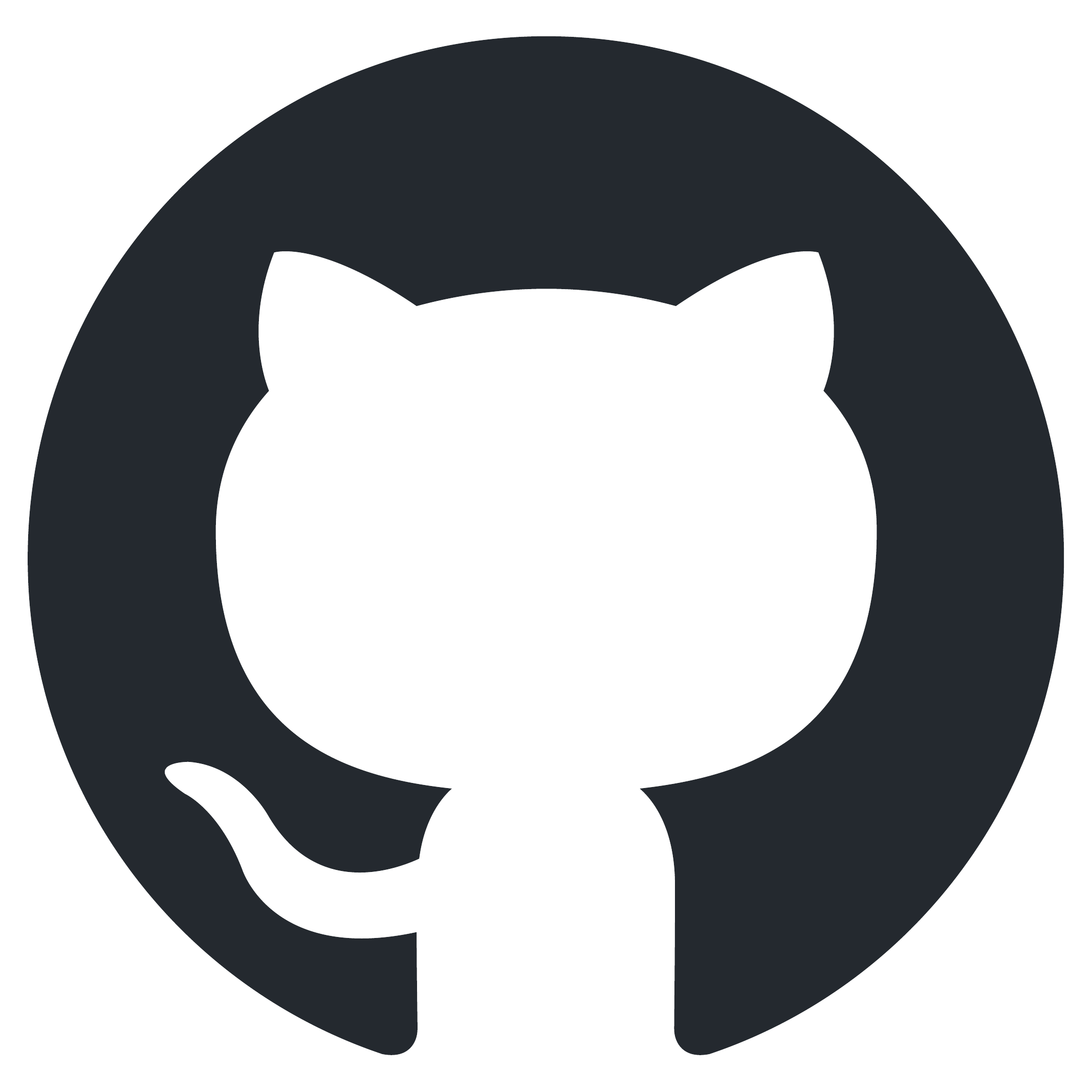}}\xspace}

\newcommand{\papername}{AiScientist}

\newtcblisting{markdownbox}[1][]{
  listing engine=listings,
  listing options={
    basicstyle=\scriptsize\ttfamily, % 保持与你原来一致的 \scriptsize
    breaklines=true,              % 自动换行
    breakatwhitespace=true,
    numbers=none,                 % 不显示行号
    columns=fullflexible,         % 优化对齐
    keepspaces=true,              % 保留代码中的空格
    aboveskip=0pt,
    belowskip=0pt
  },
  colback=gray!5!white,           % 背景色：极浅灰
  colframe=gray!70!black,         % 边框色：深灰
  title={\textbf{Prompt (Markdown)}}, % 默认标题
  fonttitle=\bfseries,
  listing only,                   % 只显示代码，不运行
  left=5mm, right=5mm, top=3mm, bottom=3mm, % 内边距
  breakable,                      % 允许跨页
  enhanced,                       % 启用高级外观
  overlay={\begin{tcbclipinterior}\fill[gray!5!white] (frame.south west) rectangle ([xshift=0mm]frame.north west);\end{tcbclipinterior}}, % 纯净背景
  #1                              % 允许修改参数
}

\newtcbinputlisting{\markdownfile}[2][]{
  listing engine=listings,
  listing file={#2}, % 读取的文件路径
  listing options={
    basicstyle=\scriptsize\ttfamily,
    breaklines=true,
    breakatwhitespace=true,
    numbers=none,
    columns=fullflexible,
    keepspaces=true
  },
  colback=gray!5!white,
  colframe=gray!70!black,
  title={\textbf{Prompt (Markdown)}},
  fonttitle=\bfseries,
  listing only,
  breakable,
  enhanced,
  #1
}

\uselogo{}

\title{Toward Autonomous Long-Horizon Engineering for ML Research}

\renewcommand{\today}{}
\newcommand{\publicday}{Apr.~15, 2026}

\author[*]{Guoxin Chen}
\author[*]{Jie Chen}

\author[ \hspace{-0.3em}]{Lei Chen}
\author[ \hspace{-0.3em}]{Jiale Zhao}
\author[ \hspace{-0.3em}]{Fanzhe Meng}
\author[$\dag$]{Wayne Xin Zhao}
\author[$\dag$]{Ruihua Song}
\author[ \hspace{-0.3em}]{Cheng Chen}
\author[ \hspace{-0.3em}]{Ji-Rong Wen}
\author[$^\dag$]{Kai Jia}

\footerlinks{%
  \github\ \href{https://github.com/AweAI-Team/AiScientist}{Github}
}

\correspondingauthor{\{gx.chen.chn, ptyzchenjie, batmanfly, jiakai0419\}@gmail.com, songruihua\_bloon@outlook.com}

\affil[1]{Gaoling School of Artificial Intelligence, Renmin University of China}
\affil[2]{Independent Researcher}
\affil[3]{AweAI Team\footnote{$^*$Equal Contributions. $^\dag$Corresponding authors.  \hfill\textbf{Date:} \publicday.}}

\begin{abstract}
Agentic systems increasingly automate pieces of AI research.
Yet turning underspecified research objectives into runnable, experimentally validated ML systems remains a central bottleneck.
We study this operational setting as \emph{long-horizon ML research engineering}: converting a research specification into a runnable ML system through repeated implementation, experimentation, and refinement.
The central challenge is to sustain cumulative project progress across heterogeneous stages under delayed, confounded feedback.
We introduce \papername{}, a multi-agent system built around \emph{thin control over thick state}: a lightweight hierarchical research team coordinates through a File-as-Bus workspace that preserves decision-relevant artifacts across roles and invocations.
On PaperBench, \papername{} improves over the strongest matched baselines by 9.92 and 11.15 points with Gemini-3-Flash and GLM-5, respectively.
On MLE-Bench Lite, it reaches 81.82 Any Medal\% under both backbones, improving over the strongest matched baselines by 4.55 and 16.67 points, and exceeding a Codex/GPT-5.5 xhigh frontier harness reference by 13.64 Any Medal points.
Ablations and process analyses show that durable project state is central to later-round refinement: removing File-as-Bus lowers PaperBench score by 6.41 points and MLE-Bench Lite Any Medal\% by 31.82 points.
These results suggest that long-horizon AI research is not only a problem of stronger local reasoning, but a systems problem of maintaining cumulative, inspectable project progress.
\end{abstract}

\makeatletter
\begin{document}

% --- hide footnote marker ONLY for title/affil footnote ---
\begingroup
\makeatletter
\renewcommand{\thefootnote}{}% no a/1/... marker
\renewcommand{\@makefnmark}{}% no mark in text
\maketitle
\makeatother
\endgroup

\begin{figure}[h] 
  \centering 
  \includegraphics[width=0.95\textwidth]{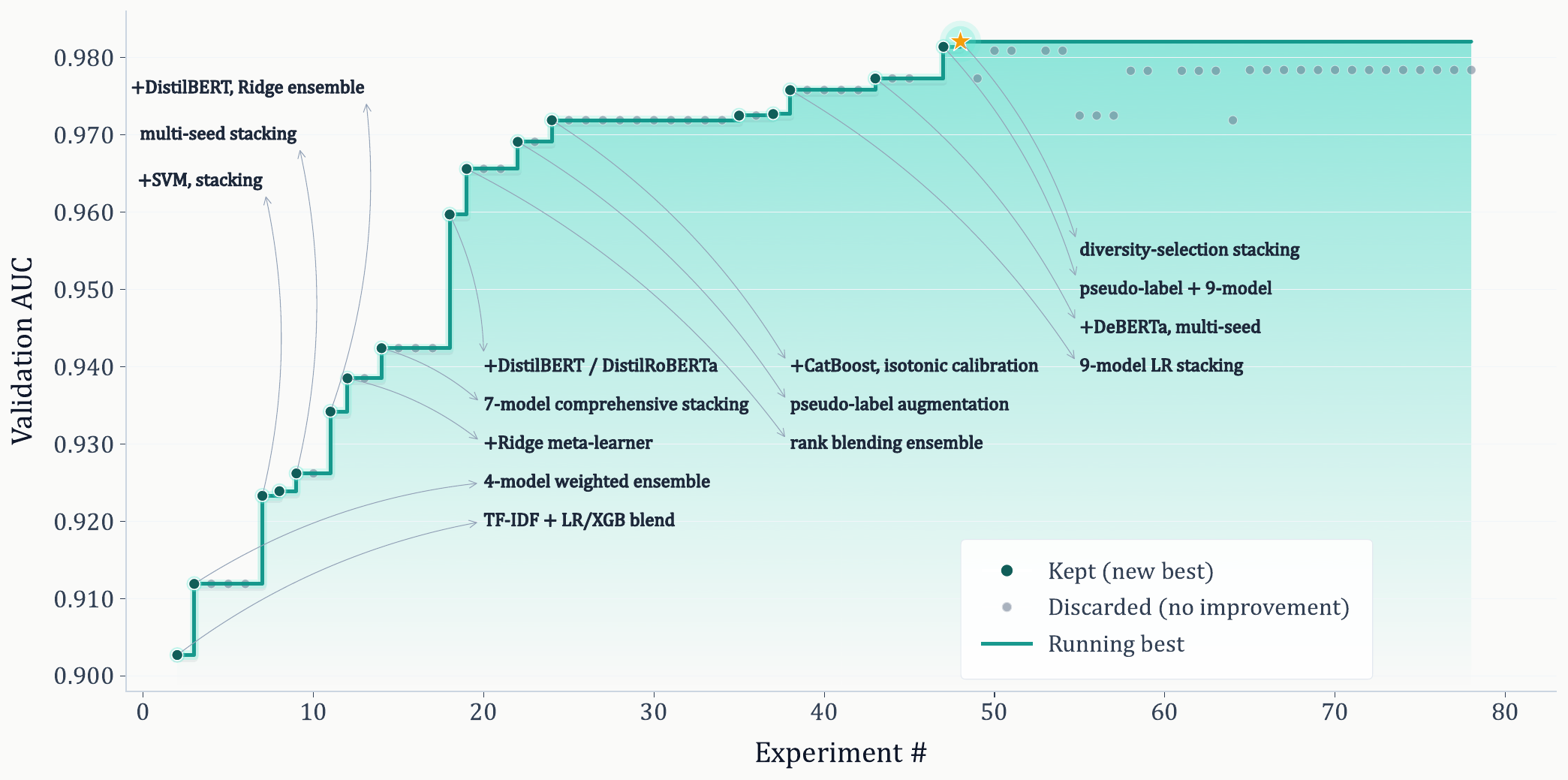} 
  \vspace{-0.3cm} 
  \caption{\papername{} autonomously improving performance on a competition-style ML task over 23 hours.
On MLE-Bench Lite's \textit{Detecting Insults} task, it conducted 74 experiment cycles without human intervention, raising validation AUC from 0.903 to 0.982 through 18 best-so-far updates.}
  \label{fig:mle-first-page}
  \vspace{-0.6cm} 
\end{figure}

\input{section/1.introduction}

\input{section/2.task}
\input{section/3.method}
\input{section/4.exp}
\input{section/5.related_work}

\input{section/6.conclusion}

\bibliography{main}
\clearpage
\appendix

\input{section/7.appendix}

% \appendix
% \newpage

% \addcontentsline{toc}{section}{Appendix}

% \begin{center}
% {\Large\bfseries\color{AweAIblue} Contents of Appendix}
% \end{center}

% \startcontents[appendix]
% \printcontents[appendix]{}{1}{\setcounter{tocdepth}{2}}

% \vspace{2em}

% \clearpage

% \input{section/8.appendix}

\end{document}

%% file: section/1.introduction.tex
\section{Introduction}
A central bottleneck in automating AI research is not only generating promising ideas, but turning underspecified research objectives into runnable, experimentally validated ML systems.
Recent agentic systems have made rapid progress on research assistance and automation, including idea generation, literature synthesis, code generation, targeted experimentation, and scientific writing~\citep{lu2024aiscientist,yamada2025aiscientistv2,tang2025airesearcher,schmidgall2025agentlaboratory,xu2026idea2story}.
Yet much of the practical difficulty of AI research lies in the engineering loop that connects these pieces: interpreting a research specification, constructing a working implementation, setting up data and environments, running experiments, diagnosing failures, and refining the system until empirical evidence supports the intended claim.
We study this operational setting as \emph{long-horizon ML research engineering}.

This setting is difficult because progress is cumulative under delayed feedback.
A paper or research specification rarely determines all implementation choices.
Important details may be implicit, missing, or scattered across sections, datasets and pretrained models must be located and integrated, and experimental discrepancies often appear only after hours of implementation and execution.
When a run fails or a metric does not match the target, the cause may lie in paper interpretation, data processing, model design, training configuration, infrastructure, or some interaction among them.
Thus the challenge is not simply to solve a sequence of local engineering subproblems, but to preserve enough evolving project state for later decisions to remain coherent.
This difficulty is visible in rigorous evaluation: on PaperBench, the best reported agent achieves only 21\% of the replication rubric, compared with 41\% achieved by top ML PhDs under a 48-hour budget~\citep{starace2025paperbench}.

The core systems problem is that agent invocations are transient, while research progress must be cumulative and persistent.
This creates two continuity requirements.
First, \emph{role continuity}: when a specialist is invoked again later, it must be able to resume from prior role-level work, including what earlier invocations tried, which assumptions they made, which failures they observed, and what questions they left open.
Second, \emph{project continuity}: different roles must coordinate around shared evidence produced across the whole project, rather than around lossy conversational handoffs.
Existing agent organizations struggle with these requirements in complementary ways.
Single-agent systems must carry an expanding history of analysis, code changes, logs, and diagnostic reasoning in a limited active context.
Multi-agent systems can distribute work, but often transfer state through compressed handoffs that may omit the detailed evidence needed for later debugging and refinement~\citep{cemri2025multiagentfail,yan2025beyondselftalk}.
For long-horizon ML research engineering, the key question is therefore how to convert \emph{transient agent activity} into \emph{durable, inspectable project progress}.

We introduce \textbf{\papername{}}, a system for autonomous long-horizon engineering for ML research designed to satisfy these two continuity requirements through \emph{thin control over thick state}.
The top-level Orchestrator provides thin control: it manages stage-level progress through concise summaries, high-level directives, and a compact workspace map, without carrying the full project history in its active context.
The shared workspace provides thick state through a \textbf{File-as-Bus protocol}: decision-relevant state is externalized into persistent project artifacts rather than left in transient conversations.
This artifact substrate supports \emph{role continuity} because a newly invoked specialist can inspect its role-owned artifacts, such as implementation rationales or experimental notes, to resume prior work.
It supports \emph{project continuity} because specialists ground their decisions in the same system of record, including paper analyses, plans, experiment records, failure diagnoses, and result summaries.
A \textbf{hierarchical research team} ties these pieces together: the Orchestrator delegates to specialized agents for paper comprehension, task prioritization, implementation, and experimentation, while durable artifacts carry detailed state across invocations and roles.

We evaluate whether this design yields sustained empirical progress on two complementary benchmarks: PaperBench~\citep{starace2025paperbench} for from-scratch paper replication and MLE-Bench Lite~\citep{chan2025mlebench} for iterative competition-style ML improvement.
Across matched comparisons, \papername{} improves over the strongest baselines by 9.92/11.15 points on PaperBench with Gemini-3-Flash/GLM-5, and reaches 81.82 Any Medal\% on MLE-Bench Lite under both backbones, improving by 4.55/16.67 points.
To calibrate these results against a frontier harness, we also report Codex~\citep{openai2025codexupgrades} with GPT-5.5 xhigh as a reference.
The best \papername{} exceeds it by 4.28 points on PaperBench and 13.64 Any Medal points on MLE-Bench Lite.
Ablations support the continuity claim: removing File-as-Bus lowers PaperBench score by 6.41 points and MLE-Bench Lite Any Medal\% by 31.82 points, with the largest degradation appearing in later-round refinement rather than first-pass executability.

To summarize, our contributions are as follows:
\begin{itemize}[leftmargin=*] %[topsep=1pt, partopsep=1pt, leftmargin=*, itemsep=-1pt]
\item We formulate long-horizon ML research engineering as a cumulative project-state problem, identifying \emph{role continuity} and \emph{project continuity} as central requirements for sustained progress under delayed experimental feedback.
\item We present \papername{}, a multi-agent system built around \emph{thin control over thick state}, where lightweight orchestration is paired with a File-as-Bus protocol that externalizes decision-relevant state into durable project artifacts.
\item We provide empirical evidence on PaperBench and MLE-Bench Lite, including matched baselines, a frontier harness reference, and ablations showing that durable state continuity is a key bottleneck for long-horizon research engineering.
\end{itemize}

%% file: section/2.task.tex
\section{Task Formulation}
\label{sec:task}
We formulate \emph{long-horizon ML research engineering} as the task of turning a research specification into a runnable ML system.
Given a research specification, an environment, a resource-access policy, and a time budget, the agent must produce a submission.
Evaluation is performed by fresh execution in a clean environment, measuring both executability and whether its empirical behavior satisfies the target objective.
This task is challenging along four dimensions:
\begin{itemize}[leftmargin=*] % [topsep=1pt, partopsep=1pt, leftmargin=*, itemsep=-1pt]
\item \textbf{Underspecification:} 
The research specification is typically underspecified rather than a complete blueprint.
Implementation details may be implicit, scattered across sections, or omitted entirely, so the agent must recover missing decisions from incomplete specifications, related literature, and other permitted public resources.
\item \textbf{System Setup Burden:}
Success depends on substantial system setup beyond code alone, including configuring environments, acquiring datasets and models from permitted sources, and integrating these resources into a runnable system.
\item \textbf{Delayed Feedback:} 
Meaningful evidence arrives only after experiments run, and discrepancies may stem from interpretation, implementation, data processing, or infrastructure. The agent must reason from delayed and often confounded feedback before deciding what to fix next.
\item \textbf{State Continuity:}
Each round of implementation and experimentation produces code, configurations, logs, results, and diagnostic evidence that later decisions must correctly interpret and build on. Progress depends on maintaining continuity across heterogeneous stages over long horizons.
\end{itemize}
\clearpage

%% file: section/3.method.tex
\section{\papername{}}
\begin{figure*}[t]
    \centering
    \includegraphics[width=\linewidth]{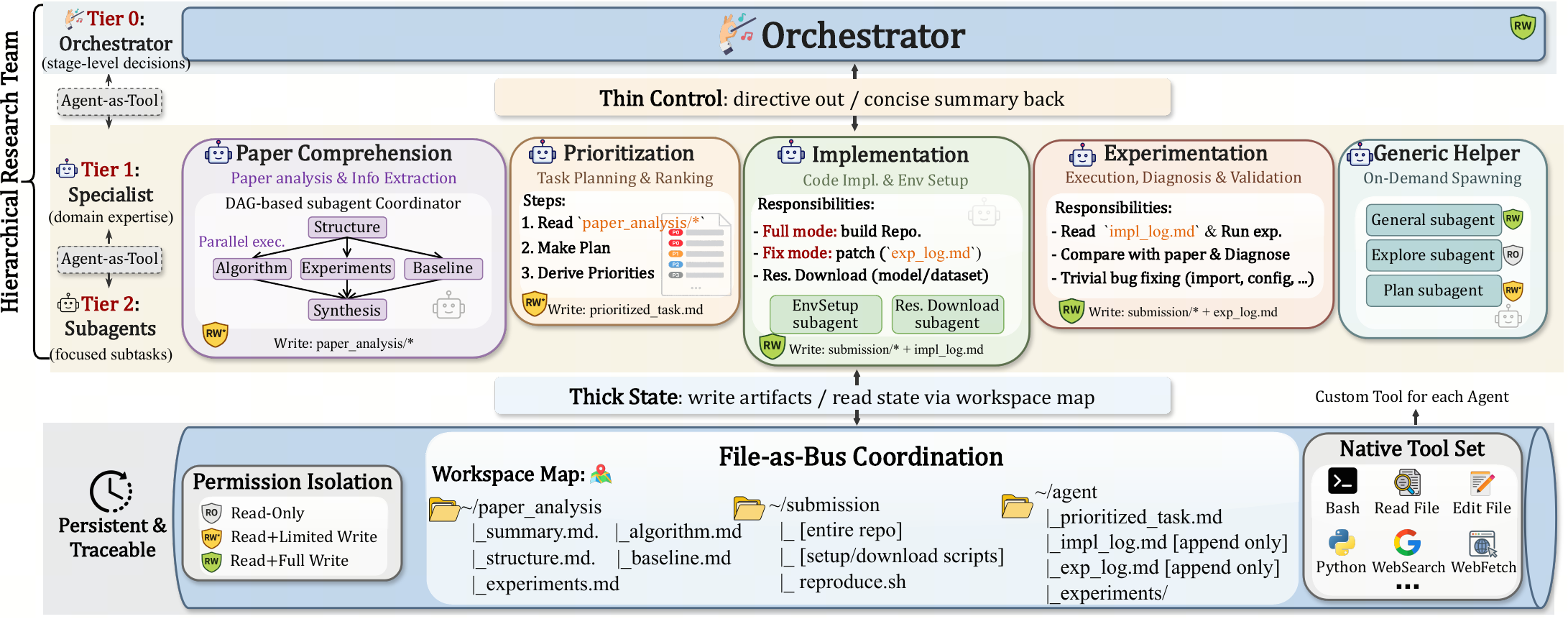}
    % \vspace{-0.8cm}
    \caption{Overview of \papername{}.
    A top-level Orchestrator maintains thin control through stage-level directives, concise summaries, and a compact workspace map.
    Specialized agents perform paper comprehension, task prioritization, implementation, experimentation, and auxiliary exploration, while coordinating through a File-as-Bus workspace that serves as the system of record.
    This design turns transient agent invocations into durable role-level and project-level progress by externalizing decision-relevant state into persistent project artifacts.}
    \label{fig:framework}
\end{figure*}

\subsection{Overview: Thin Control over Thick State}
\papername{} is built to sustain progress when agent invocations are transient but research progress must remain cumulative.
It operationalizes \emph{thin control over thick state} through two coupled mechanisms: a File-as-Bus protocol for durable state continuity (\S\ref{sec:file-as-bus}) and a hierarchical research team for lightweight stage-level control (\S\ref{sec:hierarchical-team}).
As shown in Figure~\ref{fig:framework}, these mechanisms target both \emph{role continuity}, where later invocations of a specialist resume prior role-level work, and \emph{project continuity}, where different specialists coordinate around shared evidence rather than lossy conversational handoffs.

The coupling is important.
File-as-Bus makes detailed project state durable and inspectable, but it does not by itself decide which evidence matters next or which role should act.
The hierarchical research team provides that control, but keeps it thin by routing decisions through compact summaries and the workspace map rather than the full project history.
Together, the two mechanisms let \papername{} preserve detailed state without forcing every control decision to carry it.

\subsection{File-as-Bus Coordination}
\label{sec:file-as-bus}
\papername{} implements thick state through a File-as-Bus protocol.
File-as-Bus is not merely shared storage, but an artifact-mediated coordination protocol.
Agents coordinate by publishing and consuming durable artifacts, so that project evidence, decisions, and execution state remain inspectable across invocations.
The workspace therefore acts as the system of record for project progress.

Let $\sigma$ denote the File-as-Bus schema, which specifies each artifact's purpose, update mode, permitted writers and readers.
Let $W_t$ denote the durable workspace state at step $t$.
The workspace map is a compact runtime view of both the current artifact state and the schema:
\begin{equation}
m_t = \mathcal{M}(W_t; \sigma),
\end{equation}
where $\mathcal{M}$ combines the current artifact state in $W_t$ with schema metadata from $\sigma$ to produce a compact index of available artifacts and their roles.
Agents start from $m_t$ to identify relevant artifacts, then inspect task-specific parts of $W_t$ on demand.

The schema $\sigma$ distinguishes three update modes to match different lifecycles:
\emph{append-only logs} preserve chronological evidence such as implementation logs and experiment diagnoses;
\emph{versioned artifacts} maintain a canonical current state while retaining prior revisions, such as plans, analyses, and reports;
and \emph{mutable state} stores short-lived control information that does not require history.
This distinction matters for long-horizon work: some evidence should never be overwritten, some artifacts need a stable current version with recoverable history, and some state should remain lightweight.

Together, the schema and workspace map support \emph{role continuity} by giving each specialist persistent access to role-relevant artifacts: later invocations can inspect prior notes, unresolved blockers, and role-specific logs before acting.
They also support \emph{project continuity} by making cross-role dependencies explicit: paper comprehension informs prioritization, prioritization constrains implementation, implementation produces executable artifacts, and experimentation writes back evidence for later refinement.
In this sense, File-as-Bus is both a durable state substrate and a coordination channel.

\subsection{Hierarchical Research Team}
\label{sec:hierarchical-team}
File-as-Bus determines how project state is preserved and exposed, and the hierarchical research team determines how work is routed.
At the top level, the Orchestrator maintains a compact control context, monitors stage-level progress, and decides which bottleneck to address next.
Its outputs are concise directives to specialists rather than detailed procedural scripts, leaving each specialist to expand into the local context needed for its stage.

The key control abstraction is \emph{Agent-as-Tool}.
The Orchestrator treats specialist invocation as part of the same action space as ordinary tool use:
\begin{equation}
a_t = \pi_0(c_t, m_t), \qquad a_t \in \mathcal{T}_0 \cup \mathcal{A}_1,
\end{equation}
where $c_t$ is the Orchestrator's control context, $\mathcal{T}_0$ is its native tool set, and $\mathcal{A}_1$ is the set of Tier-1 specialists.
This makes delegation selective: the Orchestrator can perform lightweight operations directly and invoke a specialist when the next step requires focused expertise or a longer local horizon.

When specialist $\pi_j$ is invoked with directive $d_t$, it receives the directive and workspace map, reads task-relevant artifacts from $W_t$, and writes back both a concise summary and workspace updates:
\begin{align}
(s_t, \Delta W_t) &= \pi_j(d_t, m_t; W_t), \\
W_{t+1} &= \mathcal{U}(W_t, \Delta W_t).
\end{align}
Here, $\mathcal{U}$ applies the specialist's artifact updates to the File-as-Bus workspace.
The summary $s_t$ updates the Orchestrator's compact control view, while $\Delta W_t$ preserves detailed progress in the workspace.
This separation lets specialist work remain rich locally without forcing the Orchestrator to carry the full reasoning trace.

The Tier-1 specialists align with the major stages of ML research engineering:
\begin{itemize}[topsep=1pt, partopsep=1pt, leftmargin=*, itemsep=-1pt]
\item \textbf{Paper Comprehension Specialist}: extracts implementation-relevant details, target metrics, proposed methods, baselines, ambiguities, and assumptions from the research specification.
\item \textbf{Prioritization Specialist}: converts paper understanding into an ordered execution plan, ranking tasks by dependency, impact, and feasibility.
\item \textbf{Implementation Specialist}: builds or patches the system, handles setup and resource integration, and records major code-side decisions.
\item \textbf{Experimentation Specialist}: executes the pipeline, compares outcomes against target objectives, and records results, failures, and diagnoses.
\item \textbf{Generic Helper Interface}: supports focused auxiliary work such as resource lookup, exploration, or local planning.
\end{itemize}
Specialists may use tightly scoped Tier-2 subagents for bounded subtasks.
The hierarchy is not recursive: Tier-2 outputs are folded back into the invoking specialist's local context and then written to durable artifacts when they matter for later stages.

\subsection{Evidence-Driven Research-Engineering Loop}
\label{sec:evidence-loop}
\papername{} uses the mechanisms above to run an adaptive research-engineering loop rather than a rigid one-pass pipeline.
Early stages turn an underspecified research objective into an execution plan and a runnable scaffold: code, configuration, setup path, resource acquisition process, and an entry point that can be repeatedly extended and evaluated.
Once this scaffold exists, progress is driven by alternating implementation and experimentation, with each run producing executable evidence such as failures, partial successes, metric gaps, bottlenecks, and result discrepancies.
The Orchestrator uses the workspace map and recorded evidence to choose the next intervention, such as patching implementation, revising data processing or configuration, rerunning experiments, or returning to earlier analysis when an assumption is invalidated.

In this way, delayed feedback becomes cumulative progress.
Once failures are recorded as durable artifacts, they become inspectable project evidence rather than isolated local errors.
Successful changes are likewise preserved as runnable code, setup state, logs, and result records.
Later invocations can continue from prior evidence instead of rediscovering it, enabling long-horizon refinement under finite time budgets.

%% file: section/4.exp.tex
\section{Experiments}
\input{tables/paper_bench}

\subsection{Experimental Setup}
\textbf{Benchmarks.}
We evaluate \papername{} in two complementary long-horizon ML research engineering benchmarks.
\textbf{PaperBench}~\citep{starace2025paperbench} evaluates from-scratch replication of top-tier ML papers.
\textbf{MLE-Bench Lite}~\citep{chan2025mlebench} evaluates sustained improvement on competition-style ML tasks, with \textit{Any Medal\%} as the primary metric.
Together, these benchmarks test whether an agent can maintain coherent progress across heterogeneous stages under realistic time budgets, rather than succeed only in a single narrow setting.

\textbf{Baselines.}
On PaperBench, we compare against BasicAgent and IterativeAgent~\citep{starace2025paperbench} under the same evaluation protocol.
On MLE-Bench Lite, we report controlled comparisons against strong autonomous ML engineering systems with diverse designs, including AIDE~\citep{jiang2025aide}, LoongFlow~\citep{wan2025loongflow}, and ML-Master 2.0~\citep{zhu2026cognitive}.
We also include two contextual reference sets that are not treated as matched baselines: official MLE-Bench Lite leaderboard results~\citep{yang2025rdagent,li2025fm,liu2025ml,toledo2025ai,nadafian2026kapso,chen2026mars,zhang2026aibuildai}, and a frontier Codex/GPT-5.5 xhigh harness~\citep{openai2025codexupgrades} evaluated under our setup.

\textbf{Implementation Details.}
We instantiate \papername{} with two backbone LLMs, Gemini-3-Flash~\citep{googledeepmind2026gemini3flash} and GLM-5~\citep{zeng2026glm}.
Across both benchmarks, each run is allocated one H20 GPU and a 24-hour budget per task, matching the standard setting.
For PaperBench full evaluation, we follow the official evaluation protocol~\citep{starace2025paperbench} and use GPT-5.4~\citep{openai2026gpt54} as the grading model.
Under this grading setup, a full 20-task PaperBench evaluation costs approximately \$832, which materially limits large-scale repeated evaluation.
For MLE-Bench Lite, we follow the MLE-Bench convention and report mean $\pm$ SEM over 3 runs/seeds.

\input{tables/mle_main_results}
\subsection{Main Results on PaperBench}
Table~\ref{tab:paperbench_main} reports the full PaperBench evaluation.
Across both backbones, \papername{} consistently improves over the strongest matched baseline: by 9.92 points with Gemini-3-Flash and by 11.15 points with GLM-5.
These gains are obtained at substantially lower cost than IterativeAgent: \$15.67 versus \$27.44 per task under Gemini-3-Flash, and \$12.20 versus \$54.90 under GLM-5.
The comparison is informative because IterativeAgent already increases interaction relative to BasicAgent, yet remains well below \papername{} while spending more.
This suggests that long-horizon performance is not explained by more rounds alone.
Those rounds must preserve and reuse prior project evidence.
As an additional contextual reference, the best \papername{} backbone exceeds the Codex harness by 4.28 points on PaperBench average score.
\vspace{-0.5em}
% \textbf{Takeaway: More interaction alone is not enough; additional rounds help only when they build on prior progress.}

\subsection{Main Results on MLE-Bench Lite}
Table~\ref{tab:mle_bench_lite} reports MLE-Bench Lite results.
In controlled comparisons, \papername{} reaches 81.82 Any Medal\% under both backbones, improving over the strongest matched baseline by 4.55 points with Gemini-3-Flash and 16.67 points with GLM-5.
The gains are also reflected in Above Median\%, where \papername{} improves by 9.09 points under both backbones.
These results indicate that \papername{} not only produces valid submissions, but also sustains the iterative improvement needed to obtain competitive outcomes.
Relative to contextual references, \papername{} exceeds the Codex/GPT-5.5 xhigh harness by 13.64 Any Medal points and surpasses the strongest official leaderboard Any Medal result reported in Table~\ref{tab:mle_bench_lite}.

\begin{figure*}[t]
    \centering
    \begin{minipage}[t]{0.32\textwidth}
        \centering
        \includegraphics[width=\linewidth]{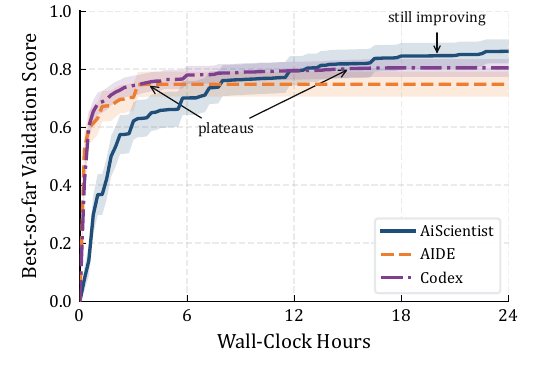}
        % \vspace{-0.25cm}
        \centerline{\small (a) Mean best-so-far score}
    \end{minipage}
    \hfill
    \begin{minipage}[t]{0.32\textwidth}
        \centering
        \includegraphics[width=\linewidth]{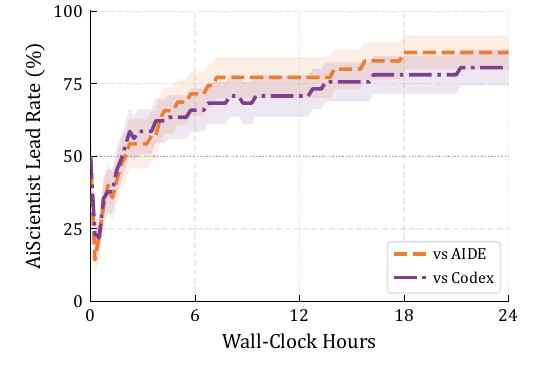}
        % \vspace{-0.25cm}
        \centerline{\small (b) Pairwise lead rate}
    \end{minipage}
    \hfill
    \begin{minipage}[t]{0.32\textwidth}
        \centering
        \includegraphics[width=\linewidth]{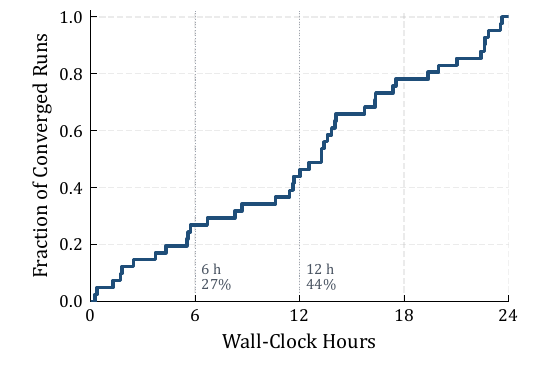}
        % \vspace{-0.25cm}
        \centerline{\small (c) Time to final best}
    \end{minipage}
    \caption{Long-horizon improvement dynamics on MLE-Bench Lite under GLM-5.
    (a) Mean validation best-so-far normalized score with $\pm 1$ SEM across task--seed trajectories.
    (b) Pairwise lead rate of \papername{} against AIDE and Codex, with binomial standard error.
    (c) Cumulative distribution of when runs first reach their final best score.
    \papername{} starts slower but continues improving after both references largely plateau. Its advantage first becomes broad across paired runs, and many runs require late-budget refinement to reach their final best.}
    \label{fig:mle_long_horizon_dynamics}
\end{figure*}
\begin{figure*}[t]
    \centering
    \begin{minipage}[t]{0.28\textwidth}
        \centering
        \includegraphics[height=3.15cm]{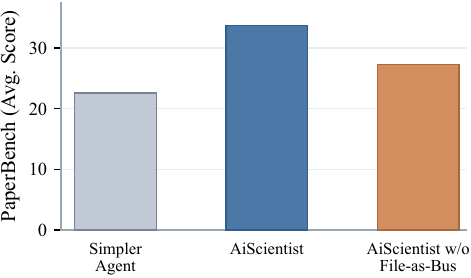}
    \end{minipage}
    \hfill
    \begin{minipage}[t]{0.68\textwidth}
        \centering
        \includegraphics[height=3.15cm]{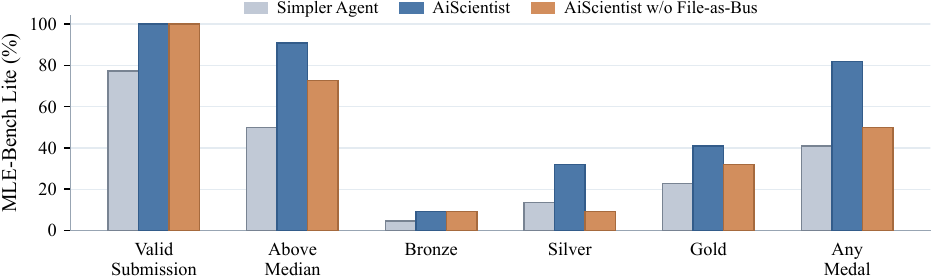}
    \end{minipage}
    % \vspace{-0.5em}
    % \caption{Mechanism analysis of \papername{} under GLM-5.
    % Left: on PaperBench, \papername{} outperforms both a simpler agent baseline and the variant without File-as-Bus.
    % Right: on MLE-Bench Lite, removing File-as-Bus leaves first-pass validity largely intact but substantially reduces stronger outcome metrics, suggesting that durable state continuity matters most for later-round refinement.}
    \caption{Mechanism analysis of \papername{} under GLM-5.
    Left: PaperBench ablations show that both hierarchical organization and File-as-Bus contribute to replication performance; removing File-as-Bus alone lowers score by 6.41 points.
    Right: on MLE-Bench Lite, File-as-Bus removal preserves first-pass validity but substantially reduces medal-level outcomes, indicating that durable state continuity primarily supports later-round refinement.}
    \label{fig:mechanism_analysis}
\end{figure*}

\subsection{Long-Horizon Improvement Dynamics}
\label{sec:long_horizon_dynamics}
Final metrics alone obscure when progress occurs.
Figure~\ref{fig:mle_long_horizon_dynamics} separates three aspects of long-horizon improvement on MLE-Bench Lite.
The average score curve measures the magnitude of performance over time; the pairwise lead-rate curve measures how broadly \papername{} is ahead across matched task--seed comparisons; and the convergence-time curve measures when runs actually reach their final best score.
AIDE and Codex improve rapidly early in the run, with Codex especially strong in the first few hours.
By contrast, \papername{} improves more slowly at the beginning, consistent with spending early budget on task understanding, planning, and scaffold construction, but continues improving after both references largely plateau.

The pairwise and convergence views clarify why this is a long-horizon effect rather than a late gain on a few outlier tasks.
\papername{} can lead on a majority of paired comparisons before its mean score overtakes the references, because early wins may be narrow while some remaining losses are still large.
At the same time, the time-to-final-best distribution shows that many runs do not reach their final best score early.
Together, the panels show that \papername{}'s advantage gradually broadens across tasks and strengthens over the full budget.

\subsection{Mechanism Analysis}
Figure~\ref{fig:mechanism_analysis} analyzes two mechanisms behind the gains of \papername{}: durable artifact-based continuity and hierarchical agent organization.
Removing File-as-Bus causes large degradation on both benchmarks.
On PaperBench, average score drops by 6.41 points.
On MLE-Bench Lite, Any Medal drops by 31.82 points.
The MLE-Bench Lite pattern is especially diagnostic: Valid Submission and Bronze remain largely intact, while Above Median, Silver, Gold, and Any Medal degrade much more.
This suggests that File-as-Bus is less important for producing a minimally runnable starting point than for preserving evidence across later rounds of diagnosis and improvement.

Hierarchical organization also appears to matter.
Relative to simpler non-hierarchical baselines, the advantage remains substantial even when File-as-Bus is removed: on PaperBench, \papername{} without File-as-Bus still improves average score by 4.74 points over the simpler agent baseline, while on MLE-Bench Lite it improves Above Median by 22.73 points and Any Medal by 9.09 points.
Together with the IterativeAgent comparison, this indicates that the gains of \papername{} are not reducible to interaction count alone.
Long-horizon ML research engineering benefits from both durable project state and a control structure that routes heterogeneous work to specialized roles.
% \textbf{Takeaway: Durable state continuity is a key bottleneck in long-horizon ML research engineering.}
% \textbf{Takeaway: File-as-Bus matters more for later-round refinement than for establishing a minimally competitive starting point.}
% \textbf{Takeaway: Simpler agent organizations are insufficient in long-horizon ML research engineering; hierarchical orchestration appears to contribute materially alongside durable state continuity.}

\subsection{Behavioral Analysis on PaperBench}
\label{sec:paperbench_behavior}
\begin{figure}[t]
    \centering
    \includegraphics[width=0.75\linewidth]{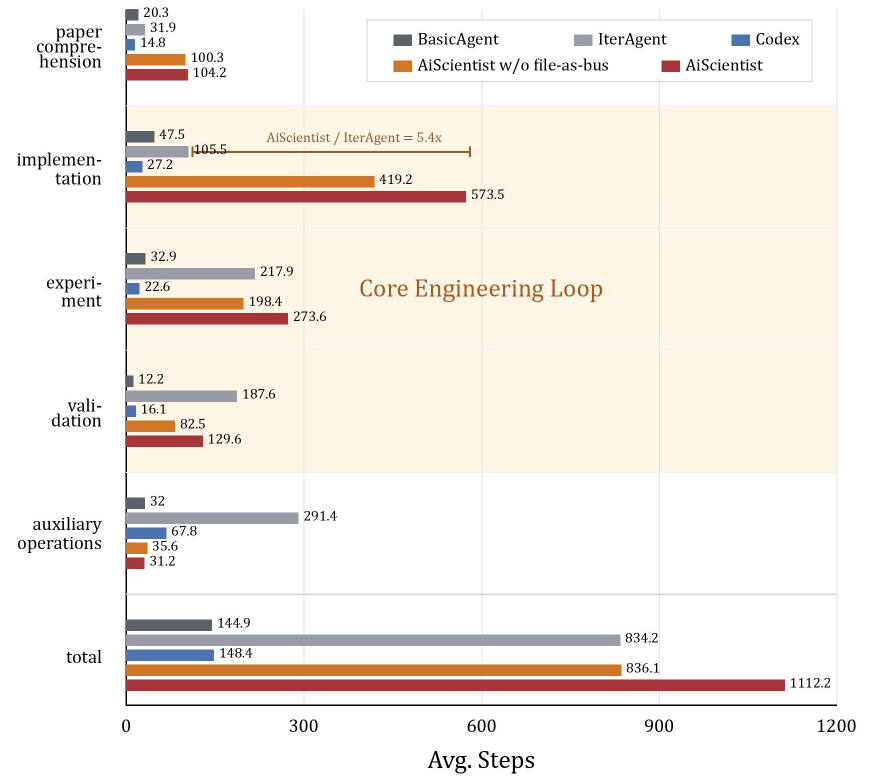}
    % \vspace{-0.8cm}
    \caption{Stage-level workflow-step allocation on PaperBench under GLM-5. Grouping steps by workflow stage enables comparison across different methods. Codex is a strong low-step reference, while \papername{} concentrates more steps in the core engineering loop.}
    \label{fig:paperbench_stage_allocation}
\end{figure}

We next analyze how different agents spend their trajectories on PaperBench.
Figure~\ref{fig:paperbench_stage_allocation} groups steps by workflow stage rather than by agent-specific roles, making the comparison meaningful across different methods.
Codex provides an important reference point: despite using far fewer steps, it remains highly competitive, reflecting the strength of both GPT-5.5 xhigh and the Codex harness.
Thus, the figure should not be read as showing that longer trajectories are intrinsically better.
The key question is where those steps are spent.

\begin{figure}[t]
    \centering
    \includegraphics[width=0.75\linewidth]{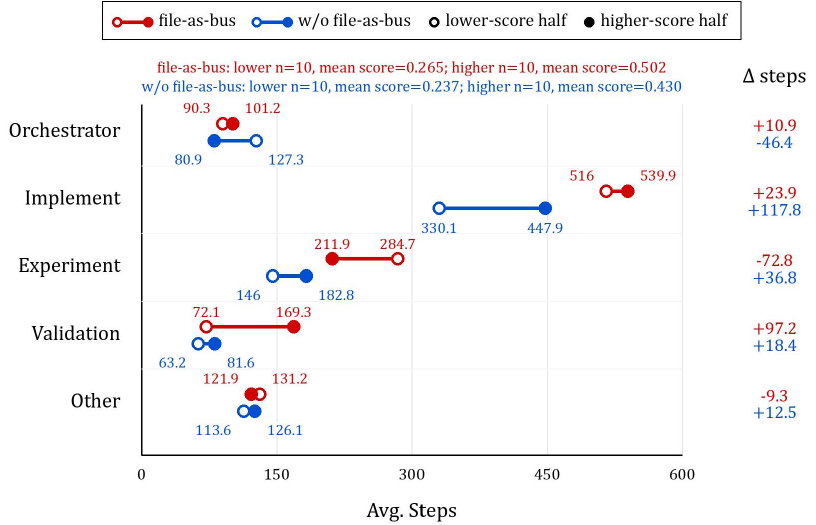}
    % \vspace{-0.7cm}
    \caption{Workflow-step allocation for higher- and lower-scoring PaperBench trajectories under GLM-5.
    Tasks are split by final PaperBench score, separately for \papername{} and its variant without File-as-Bus.
    With File-as-Bus, higher-scoring trajectories spend more of their later effort on validation and closure; without File-as-Bus, higher-scoring trajectories use substantially more implementation and experimentation steps, suggesting less efficient reuse of prior evidence.}
    \label{fig:paperbench_score_split}
\end{figure}

Compared with IterativeAgent, \papername{} shifts substantially more of its trajectory into implementation, experimentation, and validation, while IterativeAgent's many steps are less concentrated in implementation-heavy progress.
This comparison reinforces the main result: additional interaction helps only when it is organized around cumulative research-engineering work, rather than simply extending a generic agent loop.

Figure~\ref{fig:paperbench_score_split} further examines the working patterns of higher- and lower-scoring trajectories.
For \papername{}, higher-scoring trajectories are not characterized by more implementation or experimentation.
Instead, they spend more effort on validation, while using fewer experiment steps than lower-scoring trajectories.
This suggests that successful trajectories are not simply those that try more variants.
Rather, they appear to reach useful implementations earlier and spend more of the later trajectory on checking, validating, and closing the loop.

The variant without File-as-Bus shows a different pattern.
Without durable project state, higher-scoring trajectories require substantially more implementation and experimentation steps than lower-scoring trajectories.
This is consistent with the mechanism analysis in Figure~\ref{fig:mechanism_analysis}: when intermediate evidence is not preserved as effectively, success depends more heavily on repeated low-level execution effort.
With File-as-Bus, later invocations can reuse recorded assumptions, failures, experiment results, and unresolved issues, shifting successful trajectories toward evidence reuse and validation.
Together, these two behavior analyses show that \papername{}'s gains are not merely a consequence of longer trajectories, but of directing long-horizon work toward cumulative project progress.

%% file: tables/paper_bench.tex
\begin{table*}[ht]
\centering
\small 
\resizebox{0.99\linewidth}{!}{
\begin{tabular}{@{}l|c|cccc|cccc@{}}
\toprule
\multicolumn{1}{l|}{\multirow{2}{*}{\textbf{Task Name}}} & \multicolumn{1}{c|}{\makecell[c]{\textbf{GPT-5.5}}} & \multicolumn{4}{c|}{\textbf{Gemini-3-Flash}}                                                          & \multicolumn{4}{c}{\textbf{GLM-5}}                                                   \\ \cmidrule(lr){2-2} \cmidrule(lr){3-6} \cmidrule(l){7-10}
\multicolumn{1}{c|}{}                                   & \textbf{Codex}    & \textbf{BasicAgent} & \textbf{IterAgent} & \textbf{\papername{}} & \textbf{$\Delta$}                & \textbf{BasicAgent} & \textbf{IterAgent} & \textbf{\papername{}} & \textbf{$\Delta$} \\ \midrule
\multicolumn{1}{l|}{adaptive-pruning}                   & 33.42             & \underline{24.53}   & 3.05               & \textbf{27.25}        & \multicolumn{1}{c|}{\pos{2.72}}  & \underline{30.82}   & 11.93              & \textbf{33.26}        & \pos{2.44}        \\
\multicolumn{1}{l|}{all-in-one}                         & 52.93             & 20.86               & \underline{45.13}  & \textbf{46.29}        & \multicolumn{1}{c|}{\pos{1.16}}  & 33.78               & \underline{44.43}  & \textbf{49.47}        & \pos{5.04}        \\
\multicolumn{1}{l|}{bam}                                & 56.65             & \underline{48.46}   & 45.04              & \textbf{56.59}        & \multicolumn{1}{c|}{\pos{8.13}}  & \underline{51.45}   & 47.91              & \textbf{61.11}        & \pos{9.66}        \\
\multicolumn{1}{l|}{bbox}                               & 17.30             & \underline{15.43}   & 8.30               & \textbf{33.79}        & \multicolumn{1}{c|}{\pos{18.36}} & \underline{23.55}   & 19.28              & \textbf{30.02}        & \pos{6.47}        \\
\multicolumn{1}{l|}{bridging-data-gaps}                 & 14.16             & \underline{12.59}   & 12.44              & \textbf{23.09}        & \multicolumn{1}{c|}{\pos{10.50}} & 9.80                & \underline{12.50}  & \textbf{26.46}        & \pos{13.96}       \\
\multicolumn{1}{l|}{fre}                                & 14.06             & 21.67               & \underline{23.89}  & \textbf{35.21}        & \multicolumn{1}{c|}{\pos{11.32}} & \underline{21.60}   & 16.67              & \textbf{28.98}        & \pos{7.38}        \\
\multicolumn{1}{l|}{ftrl}                               & 1.50              & \underline{5.87}    & 4.15               & \textbf{10.11}        & \multicolumn{1}{c|}{\pos{4.24}}  & 3.71                & \underline{6.70}   & \textbf{8.34}         & \pos{1.64}        \\
\multicolumn{1}{l|}{lbcs}                               & 26.60             & \underline{17.75}   & 15.26              & \textbf{27.90}        & \multicolumn{1}{c|}{\pos{10.15}} & 20.68               & \underline{22.74}  & \textbf{30.10}        & \pos{7.36}        \\
\multicolumn{1}{l|}{lca-on-the-line}                    & 22.08             & 12.97               & \underline{18.30}  & \textbf{30.23}        & \multicolumn{1}{c|}{\pos{11.93}} & 22.55               & \underline{26.15}  & \textbf{28.53}        & \pos{2.38}        \\
\multicolumn{1}{l|}{mechanistic-understanding}          & 45.67             & 14.86               & \underline{21.89}  & \textbf{29.95}        & \multicolumn{1}{c|}{\pos{8.06}}  & 32.49               & \underline{34.96}  & \textbf{40.55}        & \pos{5.59}        \\
\multicolumn{1}{l|}{pinn}                               & 40.64             & 26.63               & \underline{30.81}  & \textbf{49.92}        & \multicolumn{1}{c|}{\pos{19.11}} & 22.18               & \underline{25.77}  & \textbf{58.76}        & \pos{32.99}       \\
\multicolumn{1}{l|}{rice}                               & 7.94              & \underline{10.43}   & 8.88               & \textbf{10.87}        & \multicolumn{1}{c|}{\pos{0.44}}  & \underline{6.56}    & 0.27               & \textbf{10.18}        & \pos{3.62}        \\
\multicolumn{1}{l|}{robust-clip}                        & 30.14             & \underline{15.45}   & 10.43              & \textbf{18.28}        & \multicolumn{1}{c|}{\pos{2.83}}  & 22.43               & \underline{27.56}  & \textbf{28.66}        & \pos{1.10}        \\
\multicolumn{1}{l|}{sample-specific-masks}              & 57.11             & 25.39               & \underline{33.34}  & \textbf{36.77}        & \multicolumn{1}{c|}{\pos{3.43}}  & 36.93               & \underline{41.26}  & \textbf{44.13}        & \pos{2.87}        \\
\multicolumn{1}{l|}{sapg}                               & 18.15             & 11.45               & \underline{12.65}  & \textbf{19.85}        & \multicolumn{1}{c|}{\pos{7.20}}  & \underline{6.99}    & 4.95               & \textbf{31.69}        & \pos{24.70}       \\
\multicolumn{1}{l|}{sequential-neural}                  & 41.67             & 53.51               & \underline{60.24}  & \textbf{64.94}        & \multicolumn{1}{c|}{\pos{4.70}}  & 27.2                & \underline{35.53}  & \textbf{49.32}        & \pos{13.79}       \\
\multicolumn{1}{l|}{stay-on-topic}                      & 32.31             & 8.37                & \underline{13.69}  & \textbf{20.13}        & \multicolumn{1}{c|}{\pos{6.44}}  & 3.69                & \underline{8.81}   & \textbf{14.81}        & \pos{6.00}        \\
\multicolumn{1}{l|}{stochastic-interpolants}            & 41.12             & 17.04               & \underline{17.37}  & \textbf{18.81}        & \multicolumn{1}{c|}{\pos{1.44}}  & \underline{32.18}   & 28.06              & \textbf{42.10}        & \pos{9.92}        \\
\multicolumn{1}{l|}{test-time-model-adaptation}         & 26.28             & 15.27               & \underline{18.13}  & \textbf{32.45}        & \multicolumn{1}{c|}{\pos{14.32}} & 17.81               & \underline{21.19}  & \textbf{27.33}        & \pos{6.14}        \\
\multicolumn{1}{l|}{what-will-my-model-forget}          & 9.35              & 6.61                & \underline{8.99}   & \textbf{17.87}        & \multicolumn{1}{c|}{\pos{8.88}}  & \underline{25.14}   & 10.75              & \textbf{30.82}        & \pos{5.68}        \\ \midrule
\multicolumn{1}{l|}{\textbf{Average Score}}             & 29.45             & 19.26               & \underline{20.60}  & \textbf{30.52}        & \multicolumn{1}{c|}{\pos{9.92}}  & \underline{22.58}   & 22.37              & \textbf{33.73}        & \pos{11.15}       \\
\multicolumn{1}{l|}{\textbf{Avg Cost / Task}}           & \$33.05           & \$6.25              & \$27.44            & \$15.67               & \multicolumn{1}{c|}{-}           & \$4.90              & \$54.90            & \$12.20               & -                 \\ \bottomrule
\end{tabular}
}
% \vspace{-0.5em}
\caption{Main results on PaperBench full evaluation.
Codex with GPT-5.5 xhigh is shown as a frontier harness reference; red values indicate \papername{}'s gains over the strongest matched baseline.
\textbf{Bold} and \underline{underlined} denote the best and second-best results within each backbone block.}
\label{tab:paperbench_main} 
\end{table*}

%% file: tables/mle_main_results.tex
% end deprecated single-point table

\begin{table*}[t]
\centering
\footnotesize
\setlength{\tabcolsep}{3.5pt}
\renewcommand{\arraystretch}{1.05}
% Fit to column width (two-column: \textwidth is full span of table*)
\resizebox{1.0\textwidth}{!}{%
\begin{tabular}{@{}llcccccc@{}}
\toprule
\textbf{Agent} & \textbf{Model} &
\makecell[c]{\textbf{Valid}\\\textbf{Submission}} &
\makecell[c]{\textbf{Above}\\\textbf{Median}} &
\textbf{Bronze} & \textbf{Silver} & \textbf{Gold} &
\makecell[c]{\textbf{Any}\\\textbf{Medal}} \\
\midrule
\multicolumn{8}{c}{\cellcolor{CornflowerBlue!15}\textit{\textbf{Official MLE-Bench Leaderboard Results}}} \\
\midrule
InternAgent & Deepseek-R1 & \msem{$100.00$}{0.00} & \msem{$78.79$}{5.46} & \msem{$10.61$}{1.52} & \msem{$16.67$}{3.03} & \msem{$34.85$}{1.52} & \msem{$62.12$}{3.03} \\
ML-Master & Deepseek-R1 & \msem{$100.00$}{0.00} & \msem{$74.24$}{1.52} & \msem{$4.55$}{2.62} & \msem{$13.64$}{0.00} & \msem{$30.30$}{3.03} & \msem{$48.48$}{1.52} \\
AIRA-dojo & o3 & \msem{$100.00$}{0.00} & \msem{$70.45$}{1.60} & \msem{$7.95$}{0.86} & \msem{$12.73$}{1.42} & \msem{$34.32$}{1.02} & \msem{$55.00$}{1.47} \\
ML-Master 2.0 & Deepseek-V3.2-Spe & \msem{$100.00$}{0.00} & \msem{$84.85$}{1.52} & \msem{$13.64$}{2.62} & \msem{$31.82$}{5.25} & \msem{$30.30$}{3.03} & \msem{$75.76$}{1.52} \\
R\&D-Agent & GPT-5 & \msem{$77.27$}{0.00} & \msem{$74.24$}{1.52} & \msem{$12.12$}{4.01} & \msem{$22.73$}{0.00} & \msem{$33.33$}{3.03} & \msem{$68.18$}{2.62} \\
Famou-Agent 2.0 & Gemini-2.5-Pro & \msem{$100.00$}{0.00} & \msem{$86.36$}{2.62} & \msem{$15.15$}{4.01} & \msem{$19.70$}{4.01} & \msem{$40.91$}{2.62} & \msem{$75.76$}{1.52} \\
MARS & Gemini-3-Pro & \msem{$100.00$}{0.00} & \msem{$89.39$}{1.52} & \msem{$6.06$}{1.52} & \msem{$15.15$}{1.52} & \msem{$53.03$}{1.52} & \msem{$74.24$}{1.52} \\
Leeroo & Gemini-3-Pro & \msem{$68.18$}{2.62} & \msem{$68.18$}{2.62} & \msem{$18.18$}{2.62} & \msem{$19.70$}{4.01} & \msem{$30.30$}{1.52} & \msem{$68.18$}{2.62} \\
AIBuildAI & Claude-Opus-4.6 & \msem{$100.00$}{0.00} & \msem{$81.82$}{0.00} & \msem{$13.64$}{6.94} & \msem{$25.76$}{4.01} & \msem{$37.88$}{4.01} & \msem{\underline{$77.27$}}{0.00} \\
\midrule
\multicolumn{8}{c}{\cellcolor{CornflowerBlue!15}\textit{\textbf{Frontier Harness}}} \\
\midrule
Codex & GPT-5.5 (xhigh) & \msem{$100.00$}{0.00} & \msem{$81.82$}{0.00} & \msem{$1.52$}{1.52} & \msem{$19.70$}{1.52} & \msem{$46.97$}{4.01} & \msem{$68.18$}{2.62} \\
\midrule
\multicolumn{8}{c}{\cellcolor{CornflowerBlue!15}\textit{\textbf{Controlled Evaluation}}} \\
\midrule
AIDE & Gemini-3-Flash & \msem{$87.88$}{5.46} & \msem{$59.09$}{2.62} & \msem{$7.58$}{1.52} & \msem{$9.09$}{0.00} & \msem{$28.79$}{1.52} & \msem{$45.45$}{0.00} \\
LoongFlow & Gemini-3-Flash & \msem{$77.27$}{0.00} & \msem{$77.27$}{0.00} & \msem{$12.12$}{5.46} & \msem{$25.76$}{5.46} & \msem{$39.39$}{3.03} & \msem{\underline{$77.27$}}{0.00} \\
\textbf{\papername{} (Ours)} & Gemini-3-Flash & \msem{$100.00$}{0.00} & \msem{$86.36$}{0.00} & \msem{$16.67$}{1.52} & \msem{$25.76$}{3.03} & \msem{$39.39$}{4.01} & \msem{\textbf{81.82}}{0.00} \\
\midrule
AIDE & GLM-5 & \msem{$66.67$}{5.46} & \msem{$42.42$}{4.01} & \msem{$7.58$}{3.03} & \msem{$6.06$}{4.01} & \msem{$21.21$}{1.52} & \msem{$34.85$}{3.03} \\
ML-Master 2.0 & GLM-5 & \msem{$100.00$}{0.00} & \msem{$80.30$}{1.52} & \msem{$16.67$}{1.52} & \msem{$16.67$}{1.52} & \msem{$31.82$}{0.00} & \msem{$65.15$}{1.52} \\
\textbf{\papername{} (Ours)} & GLM-5 & \msem{$100.00$}{0.00} & \msem{$89.39$}{1.52} & \msem{$13.64$}{2.62} & \msem{$25.76$}{3.03} & \msem{$42.42$}{1.52} & \msem{\textbf{81.82}}{0.00} \\
\bottomrule
\end{tabular}%
}%
% \vspace{-0.5em}
\caption{Main results on MLE-Bench Lite.
Values are percentages reported as mean $\pm$ SEM over three runs/seeds following the MLE-Bench convention.
Official leaderboard rows are contextual references, the Codex/GPT-5.5 xhigh row is a frontier harness reference evaluated under our setup, and controlled evaluation rows are matched comparisons.
\textbf{Bold} and \underline{underlined} denote the best and second-best \textit{Any Medal} performance.}
\label{tab:mle_bench_lite}
\end{table*}

%% file: section/5.related_work.tex
\section{Related Work}
\label{sec:related_work}
Recent systems have advanced AI research across scientific discovery~\citep{schmidgall2025agentlaboratory,tang2025airesearcher}, objective-driven ML engineering~\citep{jiang2025aide,zhu2026cognitive}, and paper-to-code reproduction~\citep{zhou2025repro,seo2026paper2code}.
These systems show that agents can contribute to different parts of research.
We build on this progress and study a complementary question: how to sustain coherent research-engineering progress over long horizons.
\papername{} addresses this by combining hierarchical orchestration with artifact-mediated project state, enabling specialists to coordinate through durable evidence rather than transient conversational handoffs.

%% file: section/6.conclusion.tex
\section{Conclusion}
\label{sec:conclusion}
We study long-horizon ML research engineering, where agents must turn underspecified research objectives into runnable systems and sustain progress under delayed, confounded experimental feedback.
\papername{} addresses this setting through thin control over thick state: hierarchical orchestration over a File-as-Bus workspace that preserves decision-relevant artifacts across roles and invocations.
Across PaperBench and MLE-Bench Lite, \papername{} improves over strong matched baselines, exceeds a frontier Codex/GPT-5.5 harness reference on key metrics, and shows large degradation when durable project state is removed.
Together, the results suggest that long-horizon AI research automation is not only a problem of stronger local reasoning, but a systems problem of maintaining cumulative, inspectable project progress.

% We study autonomous long-horizon ML research engineering, where agents must build runnable systems and sustain progress under delayed, confounded experimental feedback.

%% file: section/7.appendix.tex
\section{Additional Related Work}
\label{app:additional_related_work}

\subsection{Automating AI Research}

Recent work has rapidly advanced AI research automation.
Broadly, this progress spans three complementary directions.
First, automated scientific discovery and research-assistant systems study how agents can generate ideas, synthesize literature, run targeted experiments, and draft scientific artifacts~\citep{lu2024aiscientist,yamada2025aiscientistv2,tang2025airesearcher,schmidgall2025agentlaboratory,schmidgall2025agentrxiv,weng2026deepscientist,analemma2026fars,liu2026autoresearchclaw}.
Second, objective-driven ML engineering agents study iterative propose--implement--evaluate loops under explicit metrics or benchmark objectives~\citep{chan2025mlebench,jiang2025aide,zhu2026cognitive,wan2025loongflow,nam2025mlestar,chen2026mars,karpathy2026autoresearch}.
Third, paper-to-code and reproduction-oriented systems focus on translating research papers into runnable repositories or improving implementation fidelity~\citep{zhou2025repro,jansen2025codescientist,li2025deepcode,seo2026paper2code,chen2026beyondswe,fu2026davinci}.
Together, these directions establish many ingredients for autonomous AI research.
\papername{} builds on this progress by focusing on a more operationally demanding systems question: how to sustain coherent research-engineering progress across paper understanding, implementation, experimentation, and refinement.

\subsection{Multi-Agent Coordination and Long-Horizon Continuity}

Multi-agent coordination is a common strategy for extending LLM-based problem solving.
Frameworks such as CAMEL, MetaGPT, and ChatDev show that role decomposition and structured collaboration can improve complex task solving~\citep{li2023camel,hong2023metagpt,qian-etal-2024-chatdev}.
More recent systems bring related coordination patterns to research-oriented and long-horizon workflows~\citep{schmidgall2025agentlaboratory,chen2025cpo,pu2025piflow,yu-etal-2025-table,wan2025loongflow,zhao2026immersion,zhu2026cognitive}.
At the same time, recent analyses suggest that multi-agent systems can fail through brittle handoffs, weak verification, and loss of decision-relevant context~\citep{cemri2025multiagentfail,yan2025beyondselftalk,tang2026agent}.
Our work treats long-horizon performance as a problem of both orchestration and continuity.
Rather than relying primarily on conversational handoffs, \papername{} externalizes paper analyses, plans, code-side decisions, and experimental evidence into durable artifacts that downstream agents can repeatedly inspect and build on.
This design directly targets the two continuity requirements introduced in the main text: \emph{role continuity}, where later invocations of a specialist resume prior role-level work, and \emph{project continuity}, where different specialists coordinate around shared evidence.
In this sense, \papername{} is not simply another hierarchical multi-agent arrangement, but a coordination design for long-horizon ML research engineering centered on artifact-mediated continuity and \emph{thin control over thick state}.

\section{File-as-Bus Implementation Details}
\label{app:file_as_bus}

File-as-Bus is implemented as a schema-governed artifact protocol rather than as an unstructured working directory, expanding the method description in Section~\ref{sec:file-as-bus} and Figure~\ref{fig:framework}.
Each artifact has a declared purpose, update mode, writer set, and reader set.
The schema gives each specialist a stable contract: which evidence it should consult before acting, which artifacts it owns, and which downstream roles will consume its outputs.
Table~\ref{tab:fab_schema} summarizes the artifacts used by \papername{}.

\input{tables/file_as_bus_schema}

The update mode determines how an artifact may evolve.
\emph{Versioned artifacts} have a canonical current version while preserving prior revisions conceptually; they are used for analyses and plans whose latest form should be easy to find, but whose history may matter for later debugging.
\emph{Append-only logs} preserve chronological evidence and should not be rewritten; they are used for implementation rationales, experiment traces, failures, and diagnoses.
\emph{Mutable runnable artifacts} are directly edited project state, such as source code, setup scripts, and execution entry points.
Their detailed decision history is preserved through the append-only logs rather than by versioning every file in the runnable repository.

The access policy is role-scoped.
Tier-1 specialists write primarily to artifacts corresponding to their responsibility, while all roles can inspect upstream evidence that affects their decisions.
Tier-2 subagents are used for bounded subtasks and default to read-only access unless the invoking specialist folds their findings back into the appropriate artifact.
After each specialist invocation, the workspace map is refreshed from the current artifact state and schema metadata, giving the Orchestrator a compact index of available evidence without loading the entire workspace into its active context.
This is the implementation-level mechanism behind the thin-control interface described in Section~\ref{sec:hierarchical-team}.

\section{Evaluation and Metric Details}
\label{app:evaluation_details}

\paragraph{Code and artifacts.}
We include the anonymized implementation, evaluation scripts, and experiment artifacts in the supplementary material submitted with this paper.

\paragraph{PaperBench.}
PaperBench~\citep{starace2025paperbench} full evaluation contains 20 paper-replication tasks.
For each task, the agent receives the paper specification, a clean execution environment, permitted external-resource access, one H20 GPU, and a 24-hour budget.
The agent must build a fresh implementation without using the authors' original code or other blacklisted resources.
We follow the official full-evaluation protocol and use GPT-5.4~\citep{openai2026gpt54} as the grading model.
Because a full 20-task evaluation under this grading setup costs approximately \$832, repeated full-benchmark evaluations are materially constrained.

\paragraph{MLE-Bench Lite.}
MLE-Bench Lite~\citep{chan2025mlebench} contains 22 competition-style ML tasks.
Each run has a 24-hour budget and one H20 GPU.
We report metrics over three runs/seeds following the MLE-Bench convention, with \textit{Any Medal\%} as the primary metric.
Controlled rows in Table~\ref{tab:mle_bench_lite} use our evaluation setup, while official leaderboard rows are contextual references reported from the MLE-Bench ecosystem.

\paragraph{Anytime score curves.}
This definition corresponds to the mean best-so-far score panel in Figure~\ref{fig:mle_long_horizon_dynamics}.
For each wall-clock time point $t$, let $x_i(t)$ be the validation best-so-far normalized score for valid task--seed trajectory $i$, and let $n_t$ be the number of valid trajectories at time $t$.
The plotted mean is
\begin{equation}
\bar{x}(t) = \frac{1}{n_t}\sum_{i=1}^{n_t} x_i(t).
\end{equation}
Let $s(t)$ be the sample standard deviation of $\{x_i(t)\}_{i=1}^{n_t}$.
The shaded band is $\bar{x}(t) \pm s(t)/\sqrt{n_t}$, i.e., $\pm 1$ standard error of the mean (SEM) across task--seed trajectories at that time point.

\paragraph{Pairwise lead rate.}
This definition corresponds to the pairwise lead-rate panel in Figure~\ref{fig:mle_long_horizon_dynamics}.
For each matched task--seed comparison at time $t$, we assign
\begin{equation}
w_i(t) =
\begin{cases}
1, & \text{if \papername{} leads the reference},\\
0.5, & \text{if tied},\\
0, & \text{if the reference leads \papername{}}.
\end{cases}
\end{equation}
The lead rate is $\hat{p}(t)=\frac{1}{n}\sum_{i=1}^{n} w_i(t)$.
For visualization, shaded bands use the binomial standard-error approximation
\begin{equation}
\sqrt{\frac{\hat{p}(t)(1-\hat{p}(t))}{n}}.
\end{equation}
This view complements the average score curve: the score curve measures the magnitude of improvement, while the lead-rate curve measures how broadly \papername{} is ahead across matched trajectories.

\section{Baseline and Reference Details}
\label{app:baseline_details}

We distinguish matched baselines from contextual references.
Matched baselines are evaluated under the same controlled setup as \papername{} and support direct system-level comparisons.
Contextual references provide calibration against strong external systems or official leaderboard results, but are not treated as matched baselines because they may differ in model, harness, or reporting protocol.

On PaperBench, BasicAgent and IterativeAgent are the matched baselines from the official benchmark protocol~\citep{starace2025paperbench}.
They are evaluated under the same task suite and grading setup as \papername{}.
Codex with GPT-5.5 xhigh is reported as a frontier harness reference~\citep{openai2025codexupgrades}: it is useful for calibrating against a strong contemporary coding setup, but it is not a matched baseline because both the harness and backbone model differ from our controlled comparisons.

On MLE-Bench Lite, AIDE~\citep{jiang2025aide}, LoongFlow~\citep{wan2025loongflow}, and ML-Master 2.0~\citep{zhu2026cognitive} are controlled comparison systems in our setup.
Official MLE-Bench Lite leaderboard rows are included as contextual references to place the controlled results in the broader ecosystem~\citep{yang2025rdagent,li2025fm,liu2025ml,toledo2025ai,nadafian2026kapso,chen2026mars,zhang2026aibuildai}.
The Codex/GPT-5.5 xhigh row is again a frontier harness reference evaluated under our setup, intended to answer how \papername{} compares with a strong black-box coding harness rather than to isolate the effect of a single architectural choice.

\section{Supplementary Analyses}
\label{app:supplementary_analyses}

\subsection{Delegation Patterns on PaperBench}
\label{app:delegation_patterns}

\begin{figure}[t]
    \centering
    \includegraphics[width=0.75\linewidth]{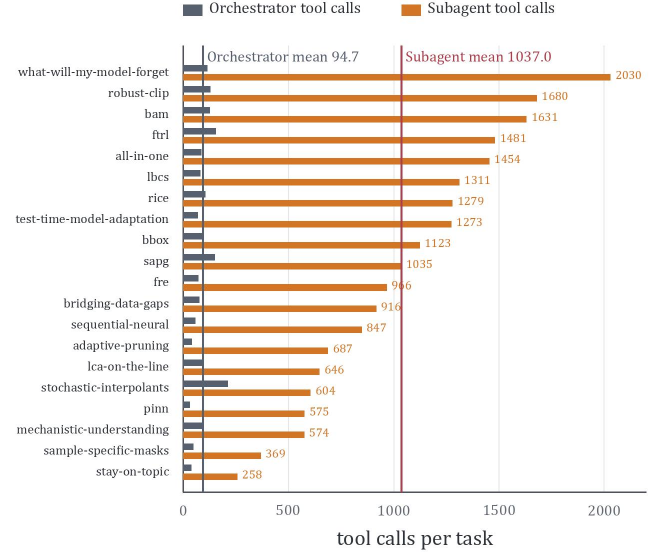}
    \caption{Delegation ratio across PaperBench tasks under GLM-5.
    Subagent delegation is not concentrated in a small number of tasks; it appears across most tasks, indicating that the hierarchical team is used as a systematic control mechanism rather than as an occasional fallback.}
    \label{fig:app_delegation_ratio}
\end{figure}

Figure~\ref{fig:app_delegation_ratio} complements the workflow-step analysis in Section~\ref{sec:paperbench_behavior}, especially Figure~\ref{fig:paperbench_stage_allocation}.
The result shows that \papername{} regularly invokes specialist and subagent work across PaperBench tasks.
This is consistent with the intended thin-control design: the Orchestrator maintains stage-level control, while substantial local work is delegated to focused agents whose outputs are folded back into the shared workspace.

\subsection{Budget and Medal Outcomes on MLE-Bench Lite}
\label{app:budget_medal}

\begin{figure}[t]
    \centering
    \includegraphics[width=0.5\linewidth]{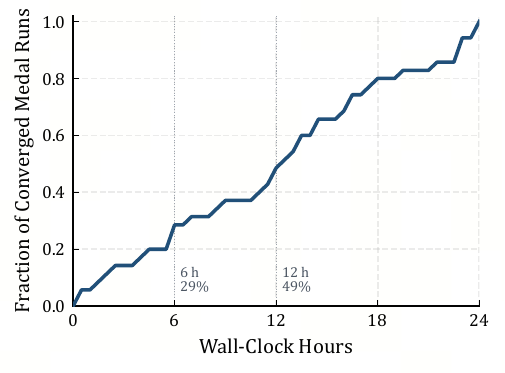}
    \caption{Budget-dependent medal outcomes on MLE-Bench Lite under GLM-5.
    The figure provides a supplementary view of how competitive outcomes emerge over the 24-hour budget.}
    \label{fig:app_budget_medal}
\end{figure}
% \begin{figure*}[t]
%     \centering
%     \includegraphics[width=\linewidth]{images/detecting_insults_progress.pdf}
%     \caption{\papername{} autonomously improves performance on a competition-style ML task over 23 hours.
% On MLE-Bench Lite's \textit{Detecting Insults} task, it conducts 74 experiment cycles without human intervention, raising validation AUC from 0.903 to 0.982 through 18 best-so-far updates.
% This case illustrates how repeated experimentation and durable experiment records support continued refinement within a single task.}
%     \label{fig:app_detecting_insults}
% \end{figure*}

Figure~\ref{fig:app_budget_medal} provides a complementary view of long-horizon behavior on MLE-Bench Lite.
Where Figure~\ref{fig:mle_long_horizon_dynamics} focuses on score trajectories, pairwise lead rates, and time to final best score, this analysis summarizes how medal-level outcomes depend on the available budget.
It supports the same interpretation as the main dynamics figure: later-budget refinement is important for converting runnable submissions into stronger competitive outcomes.

% \subsection{Case Study: Detecting Insults}
% \label{app:detecting_insults}

% Figure~\ref{fig:app_detecting_insults} illustrates the behavior behind the aggregate long-horizon trends in Figure~\ref{fig:mle_long_horizon_dynamics}.
% On a single competition-style task, \papername{} continues to improve after the initial runnable solution, using repeated experiment cycles to refine the model and validation performance.
% The case is not intended as an additional benchmark result; rather, it shows qualitatively how the evidence-driven loop operates within one task trajectory.

%% file: tables/file_as_bus_schema.tex
\begin{table*}[t]
\centering
\small
\setlength{\tabcolsep}{3.5pt}
\resizebox{\linewidth}{!}{
\begin{tabular}{@{}p{0.20\linewidth}p{0.30\linewidth}p{0.20\linewidth}p{0.20\linewidth}p{0.20\linewidth}@{}}
\toprule
\textbf{Artifact} & \textbf{Purpose} & \textbf{Update Mode} & \textbf{Writer} & \textbf{Readers} \\
\midrule
paper\_analysis/summary.md &
Concise implementation-oriented summary of the paper, including task goal, target result, and key assumptions. &
Versioned artifact &
Paper Comprehension &
Orchestrator, Prioritization, Implementation, Experimentation \\
paper\_analysis/structured.md &
Structured extraction of datasets, metrics, methods, baselines, training details, and resource requirements. &
Versioned artifact &
Paper Comprehension &
Orchestrator, Prioritization, Implementation, Experimentation \\
paper\_analysis/algorithm.md &
Algorithm-level reconstruction of the proposed method and implementation-relevant details. &
Versioned artifact &
Paper Comprehension &
Implementation, Experimentation, Orchestrator \\
paper\_analysis/baseline.md &
Baseline methods, expected comparisons, and result targets that should be reproduced or approximated. &
Versioned artifact &
Paper Comprehension &
Prioritization, Implementation, Experimentation, Orchestrator \\
agent/prioritized\_task.md &
Current ordered task plan, including dependencies, expected impact, feasibility, and next actions. &
Versioned artifact &
Prioritization &
Orchestrator, Implementation, Experimentation \\
submission/ &
Runnable repository under construction, including source code, configuration, scripts, and generated project files. &
Mutable runnable artifact &
Implementation &
Implementation, Experimentation, Orchestrator \\
submission/setup/ &
Setup and resource-acquisition scripts, including dependency installation, dataset download, and model preparation. &
Mutable runnable artifact &
Implementation &
Experimentation, Orchestrator \\
submission/reproduce.sh &
Canonical entry point used to rerun the current system in evaluation or validation. &
Mutable runnable artifact &
Implementation &
Experimentation, Orchestrator \\
agent/impl\_log.md &
Append-only record of implementation decisions, blockers, deviations from paper analysis, and unresolved code-side issues. &
Append-only log &
Implementation &
Orchestrator, Paper Comprehension, Prioritization, Experimentation \\
agent/exp\_log.md &
Append-only record of experimental runs, metrics, failures, diagnoses, and evidence used for later refinement. &
Append-only log &
Experimentation &
Orchestrator, Prioritization, Implementation, Paper Comprehension \\
\bottomrule
\end{tabular}
}
\caption{File-as-Bus artifact schema used by \papername{}.
Each artifact has an explicit purpose, update mode, writer, and reader set.
This schema turns the shared workspace into a collaboration contract rather than an unstructured scratchpad.}
\label{tab:fab_schema}
\end{table*}